\documentclass[lettersize,journal]{IEEEtran}
\usepackage{amsmath,amsfonts}
\usepackage{algorithmic}
\usepackage{algorithm}
\usepackage{array}
\usepackage[caption=false,font=normalsize,labelfont=sf,textfont=sf]{subfig}
\usepackage{textcomp}
\usepackage{stfloats}
\usepackage{url}
\usepackage{verbatim}
\usepackage{graphicx}
\usepackage{cite}
\usepackage{float}
\usepackage{amssymb}
\usepackage{booktabs}
\usepackage{xcolor}
\usepackage{hyperref}
\hypersetup{
  colorlinks=true,
  citecolor=blue,
  linkcolor=blue,
  urlcolor=blue,
  filecolor=blue
}
\let\oldcite\cite
\renewcommand{\cite}[1]{{\color{blue}\oldcite{#1}}}
\usepackage[all]{hypcap}
\usepackage{threeparttable}

\begin{document}

\title{Temporal Action Selection for Action Chunking}

\author{Yueyang Weng$^{1}$, Xiaopeng Zhang$^{1}$, Yongjin Mu$^{1}$, Yingcong Zhu$^{1}$, Yanjie Li$^{1}$
        % <-this % stops a space
\thanks{$^{1}$Guangdong Key Laboratory of Intelligent Morphing Mechanisms and Adaptive Robotics and School of Intelligence Science and Engineering, the Harbin Institute of Technology Shenzhen, China.}% <-this % stops a space
% % \thanks{Manuscript received April 19, 2021; revised August 16, 2021.}}
}

% {Weng \MakeLowercase{\textit{et al.}}: Temporal Action Selection for Action Chunking}


\maketitle

\begin{abstract}
Action chunking is a widely adopted approach in Learning from Demonstration (LfD). By modeling multi-step action chunks rather than single-step actions, action chunking significantly enhances modeling capabilities for human expert policies. However, because action chunking makes a single decision only after a complete action block has been executed, the resulting reduction in decision frequency restricts the utilization of real-time observations, impairing reactivity in dynamic or noisy environments. Existing efforts to address this issue have primarily resorted to trading off reactivity against decision consistency, without achieving both. To address this limitation, we propose a novel algorithm, Temporal Action Selection (TAS), which caches predicted action chunks from multiple timesteps and dynamically selects the optimal action through a lightweight selector network. TAS achieves balanced optimization across both reactivity and decision consistency. Experiments across multiple tasks with diverse base policy architectures show that TAS significantly improves success rates. Furthermore, integrating TAS as a base policy with residual reinforcement learning (RL) improves both training efficiency and the performance ceiling. Experiments in both simulation and physical robots confirm the method's efficacy.
\end{abstract}

% \begin{IEEEkeywords}
% Reinforcement Learning, Deep Learning in Grasping and Manipulation, Learning from Demonstration
% \end{IEEEkeywords}

%%%%%%%%%%%%%%%%%%%%%%%%%%%%%%%%%%%%%%%%%%%%%%%%%%%%%%%%%%%%%%%%%%%%%%%%%%%%%%%%
\section{Introduction}

\IEEEPARstart{L}{earning} from Demonstration (LfD) has emerged as a prominent paradigm for extending robotic manipulation capabilities\cite{10783000},\cite{LIU2025114613},\cite{lin2025learning},\cite{sun2022motion}. By learning human expert policies from limited demonstrations\cite{zhao2023learning},\cite{sim2025learning},\cite{george2023one}, LfD enables robots to autonomously execute complex long-horizon tasks without explicit environment modeling\cite{pmlr-v164-jang22a}. An effective approach within this paradigm is action chunking, where policies predict contiguous multi-step action chunks rather than single-step actions\cite{zhao2023learning},\cite{chi2023diffusionpolicy},\cite{lee2024behavior},\cite{fu2024mobile}. This approach significantly improves long-term decision consistency by extending the planning horizon of each prediction. At the same time, the reduced prediction frequency helps mitigate the compounding errors in imitation learning. Consequently, the adoption of action chunking substantially enhances the modeling capacity for human expert policies.

However, because action chunking makes a single decision only after a complete action block has been executed, the resulting reduction in decision frequency restricts the utilization of real-time observations\cite{black2025real}. In dynamic environments, during precise manipulation, or under sensor noise, the lack of timely correction based on live feedback can lead to performance degradation or even task failure\cite{liu2025bidirectional}.

Existing attempts to address this limitation exhibit notable shortcomings. A typical parameter adjustment is to maintain the prediction horizon while shortening the executed action horizon\cite{chi2023diffusionpolicy}. However, this manual tuning merely negotiates trade-offs between reactivity and decision consistency without resolving it. Temporal Ensemble\cite{zhao2023learning} applies an exponential moving average (EMA) to the actions at each timestep, which can improve motion smoothness. However, it disrupts the multimodal action distributions essential for representing diverse human strategies and provides limited improvements in reactivity. Bidirectional Decoding (BID)\cite{liu2025bidirectional} improves reactivity by scoring multiple candidates sampled at the current timestep. Nevertheless, it incurs substantial computational overhead due to repeated sampling and scoring. More fundamentally, BID does not fully leverage predictions from past observations, which limits its capacity to enhance decision consistency.

To overcome these limitations, we propose a Temporal Action Selection (TAS) algorithm that dynamically selects optimal actions from candidates with complementary advantages. By caching predicted action chunks temporally at each timestep, we construct candidate action sets. A lightweight trainable selector network, trained via reinforcement learning (RL), then dynamically selects optimal actions from these candidates. The TAS framework maintains compatibility with all action chunking policy architectures and autonomously balances reactivity and decision consistency across diverse tasks and execution phases. Furthermore, an optional coherence penalty can be integrated during training to explicitly enhance motion smoothness. Together, these mechanisms collectively elevate overall policy performance.

TAS provides a pathway for advanced policy refinement by first generating enhanced base policies. Specifically, these TAS-enhanced policies can be integrated with residual RL\cite{ankile2024imitationrefinementresidual},\cite{8794127},\cite{schoettler2020deep} in two distinct ways: serving as a superior starting point for residual RL, or being co-optimized jointly with the residual policy. This integration enhances training efficiency of residual RL, raises the final performance ceiling, and contributes to robust policy performance that is validated in real-world deployment. The key contributions of this paper are as follows:

\begin{itemize}
    \item We propose a general TAS framework for action chunking policies, achieving significant success rate improvements over diverse tasks and base policy architectures.
    \item We introduce an implicit Space-Aware selector network as one effective instantiation of TAS, which selects optimal actions from temporally cached candidates.
    \item We extend residual RL with TAS, improving both training efficiency and the performance ceiling. The experiments demonstrate these gains across simulation and real-world deployments.
\end{itemize}

\section{Related Work}
\textit{Learning from Demonstration}: In the field of robotic manipulation, Learning from Demonstration has been widely adopted as a means of acquiring robot policies from a limited number of human demonstrations. Despite achieving promising results in controlled laboratory settings\cite{zhao2023learning}, the broader deployment of LfD in real-world scenarios continues to face a set of critical challenges\cite{belkhale2023data}.

(i) Demonstrator-style variance: The demonstrations collected from different human operators often exhibit substantial inconsistencies in several aspects, including dominant hand preference, motion amplitude, and execution speed. Such inter-operator variability leads to a pronounced multimodal distribution in the action space: for a given task state or objective, the dataset may contain multiple distinct yet equally valid action trajectories corresponding to the stylistic preferences of different demonstrators\cite{zhao2023learning},\cite{lee2024behavior},\cite{mandlekar2022matters},\cite{shafiullah2022behavior}. This multimodality poses a significant learning challenge for standard imitation learning policies, which typically assume a unimodal or sharply peaked conditional action distribution.

(ii) Imperfect demonstrations: Unlike synthetic data generated in simulation, human demonstrations are inherently imperfect. They naturally contain involuntary hesitations, pauses between motion primitives, and occasionally suboptimal decision-making trajectories\cite{chi2023diffusionpolicy},\cite{belkhale2023data}. Since standard imitation learning algorithms typically treat all provided demonstrations as idealized expert executions to be reproduced, such noise is inadvertently learned as part of the target behavior, thereby degrading the quality and efficiency of the resulting policy.

(iii) Latent high-level intentions: Human behavior is often guided by macroscopic subgoals and long-term planning constraints that exist at a cognitive level. However, these latent high-level intentions are rarely explicitly observable in the low-level, short-term action sequences that constitute the typical LfD training sample\cite{zhao2023learning},\cite{mandlekar2022matters}. This semantic gap between observable actions and unobservable intent complicates the policy's ability to infer correct context for long-horizon tasks.

These challenges collectively undermine the reliability of standard imitation learning policies. In response, recent research efforts have sought to mitigate these specific limitations through a combination of strategies, including the development of enhanced network architectures\cite{zhao2023learning},\cite{10948321},\cite{brohan2023rt}, the introduction of novel training paradigms\cite{lee2024behavior},\cite{pearceimitating},\cite{florence2022implicit}, and the adoption of action chunking\cite{zhao2023learning},\cite{chi2023diffusionpolicy},\cite{bharadhwaj2024roboagent},\cite{ankile2024juicer}. Through these advancements, the field has achieved significant progress in improving the robustness of policies.

\textit{Analysis and Enhancement of Action Chunking}:
Liu et al.~\cite{liu2025bidirectional} have provided a detailed analysis of the mechanisms underlying the effectiveness of action chunking in imitation learning. Their analysis argues that the primary advantage of action chunking lies in its ability to enable the policy network to infer unobserved furture states from the limited observations available at each inference timestep. By predicting a sequence of future actions rather than a single immediate action, the network is effectively forced to reason over an extended temporal context window, which in turn facilitates more temporally consistent action prediction.

Several extensions and enhancements have been proposed to further improve the performance and applicability of action chunking. One such enhancement is Temporal Ensemble\cite{zhao2023learning}, which applies an EMA over the overlapping portions of consecutively predicted action chunks. This smoothing operation serves to improve the overall motion smoothness of the executed trajectory and mitigates the abrupt transitions that can otherwise occur at chunk boundaries. Another direction is represented by BID\cite{liu2025bidirectional}, an approach that repeatedly samples multiple candidate action sequences from the same observation at each timestep and subsequently selects the action deemed most favorable according to a scoring criterion. The repeated sampling and subsequent selection performed at each inference timestep improve reactivity and contribute to enhanced overall policy performance. Real-Time Chunking (RTC)\cite{black2025real} has reformulated the action chunk generation process as an in-filling problem. In this formulation, the actions that are scheduled for imminent execution are treated as a pre-filled context, thereby ensuring that newly generated action chunks remain temporally consistent with the immediately preceding execution plan. This approach has proven particularly effective in deployment scenarios where policy inference latency is non-negligible.

\textit{Integration of Reinforcement and Imitation Learning}: 
In robotic manipulation, combining reinforcement learning with imitation learning to leverage the complementary strengths of both paradigms has become an important research direction. 

A widely adopted approach is residual RL, which learns an additive corrective term (a residual policy) on top of a fixed, pre-trained imitation learning policy\cite{ankile2024imitationrefinementresidual},\cite{8794127},\cite{schoettler2020deep}. In this formulation, the base IL policy provides a stable prior, while the residual component refines the overall behavior through RL exploration, thereby combining the data efficiency of imitation with the optimization capability of reinforcement. Another common strategy is to fine-tune IL policies via reinforcement learning\cite{li2025reinforcement},\cite{chen2025pirl},\cite{intelligence2025pi}. Instead of keeping the base policy frozen, these methods directly update the parameters of the pre-trained model using either online or offline RL post-training to optimize an RL objective.

\section{Problem Formulation}
We consider a policy $\pi(a_{t:t+l-1} \mid s_{t-c+1:t})$ that predicts a chunk of $l$ future actions conditioned on a context window of $c$ past observations. For clarity, we denote by $a_t^{k}$ the action predicted for timestep $t+k$ using observations available at timestep $t$. Under this notation, $a_t^0$ corresponds to the immediate action to be executed at the current timestep $t$, while $a_t^{l-1}$ represents the final action in the predicted chunk.

This design introduces a fundamental trade-off between reactivity and decision consistency. The first action $a_t^0$ is generated using the most recent observations and therefore achieves optimal reactivity with respect to immediate sensory feedback. Conversely, the last action $a_t^{l-1}$ is informed by an implicitly extended temporal context window, which has been shown to enhance the modeling of latent patterns in human demonstrations and thereby improve decision consistency\cite{liu2025bidirectional}. As a result, the overall performance of the policy becomes critically dependent on the manually selected chunk length $l$. 
This dependence forces practitioners to tune $l$ according to the specific characteristics of the target environment. Longer chunk lengths are generally preferred in deterministic settings, whereas shorter lengths are necessary for tasks that involve sensor noise or dynamic uncertainties (e.g., peg-in-hole tasks). Such manual adjustment not only introduces an additional hyperparameter burden but also constrains the policy from realizing its full potential across varying conditions. Moreover, within structured tasks like FurnitureBench\cite{heo2023furniturebench}, involving sequential sub-tasks (positioning, grasping, transporting, aligning, inserting, screwing, among others), the relative importance of reactivity and decision consistency can shift substantially from one phase to the next.

To address these limitations, a natural strategy is to predict action chunks at each timestep while retaining multiple candidate actions that originate from different temporal contexts. At a given timestep $t$, these candidates are all applicable to the current decision yet are computed using observations drawn from varying historical horizons. Because each candidate embodies a distinct trade-off between reactivity and decision consistency, dynamically selecting among them offers a principled mechanism for autonomously balancing this trade-off in response to both environmental properties and task-phase transitions.

\section{Method}

\subsection{Temporal Action Selection Architecture}
\label{subsection: Temporal Action Selection Architecture}

\begin{figure*}[t]
    \centering
    \includegraphics[width=0.95\textwidth]{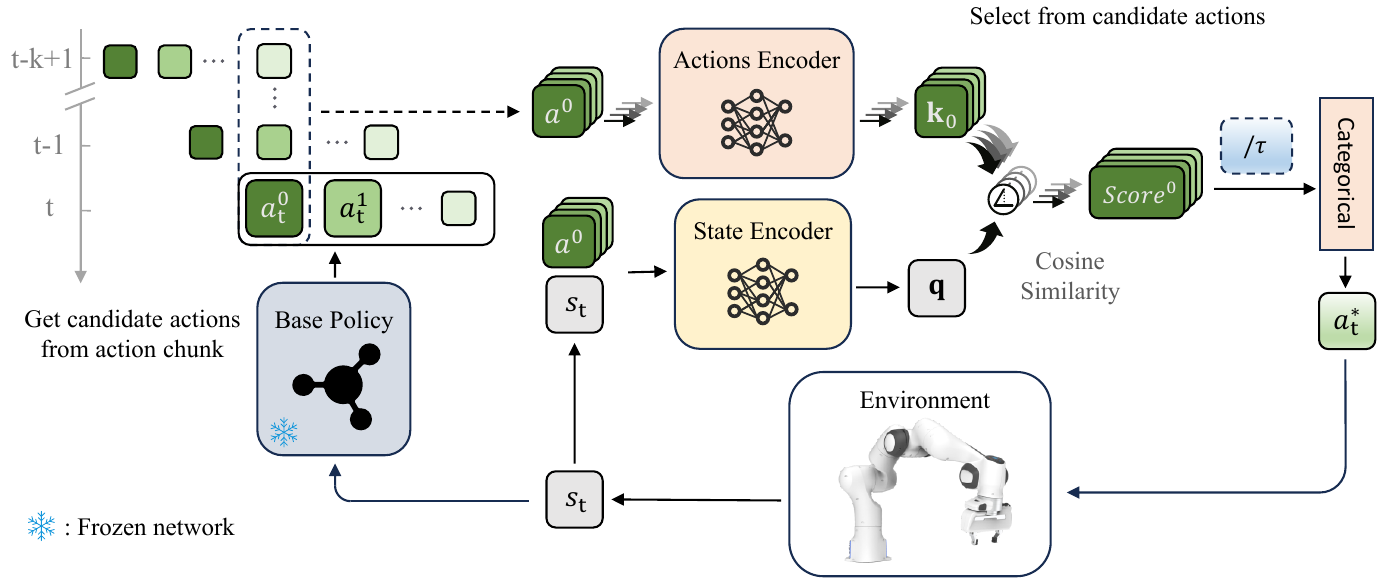}
    \caption{Overview of TAS. The base policy generates candidate actions from observations, followed by the selector network choosing optimal actions via latent-space similarity.}
    \label{fig:all}
\end{figure*}

Given a base policy $\pi(a_{t:t+l-1} \mid s_{t-c+1:t})$ that predicts action chunks of length $l$, we obtain at each timestep a set of $l$ predicted actions, each corresponding to a distinct future timestep. By caching the first $k$ predictions from the most recent $k$ timesteps (with $k \leq l$), we construct a candidate set for the current timestep $t$ as:
\[\mathcal{A}_t = \{a_t^0, a_{t-1}^1, \dots, a_{t-k+1}^{k-1}\}\]
where $a_t^0$ is the action predicted for timestep $t$ using the most recent observation $s_t$, and $a_{t-k+1}^{k-1}$ is the action predicted for timestep $t$ using the historical observation $s_{t-k+1}$. For notational brevity, we denote each candidate as $a^i$ ($i \in \{0, \dots, k-1\}$), omitting the explicit reference to its historical origin.

To select the most appropriate action from this candidate set, we introduce an implicit Space-Aware selector network inspired by the core principles of attention mechanisms\cite{vaswani2017attention}. The selector measures the relevance between the current contextual requirements of the task and each candidate action by computing cosine similarity within a shared embedding space. Architecturally, the network consists of dual encoders that correspond, respectively, to the Query and Key components of standard attention frameworks:
\begin{itemize}
    \item State encoder $\phi(\cdot)$ (produces Query): This encoder takes as input both the current observation $s_t$ and the full candidate set $\mathcal{A}_t$. It outputs a query vector $\mathbf{q}$ that encapsulates the task's immediate contextual requirements. When the observation space includes images, raw pixels are first passed through the visual encoder of the base policy to extract a suitable feature representation.
    \item Action encoder $\psi(\cdot)$ (produces Keys): This encoder processes each individual action candidate $a^i \in \mathcal{A}_t$ and produces a corresponding key vector $\mathbf{k}_i$, which characterizes the intrinsic properties of that action.
\end{itemize}
It is important to note that the state encoder receives not only $s_t$ but also the entire candidate set $\mathcal{A}_t$. Since each candidate $a^i$ is derived from observations at different historical timesteps, $\mathcal{A}_t$ serves as a compressed summary of recent temporal context. Incorporating this information allows the query vector $\mathbf{q}$ to capture richer contextual requirements beyond what is available from the instantaneous observation alone.

Action scores are computed as:
\begin{equation}
\mathrm{Score}\left(a^i\right) = \frac{\phi\left(s_t, \mathcal{A}_t\right) \cdot \psi\left(a^i\right)}{\left\| \phi\left(s_t, \mathcal{A}_t\right) \right\| \left\| \psi\left(a^i\right) \right\|}
\end{equation}
Action probabilities are derived through scaled softmax:
\begin{equation}
P\left(a^i\right) = \frac{\exp\left( \mathrm{Score}\left(a^i\right) / \tau \right)}{\sum_{j=0}^{k-1} \exp\left( \mathrm{Score}\left(a^j\right) / \tau \right)}
\end{equation}
where $\tau \in \mathbb{R}^+$ denotes the temperature coefficient controlling exploration-exploitation trade-off. The complete workflow is illustrated in Fig.~\ref{fig:all}. During training, actions are sampled from the categorical distribution $P(a^i)$ to enable exploration:
\begin{equation}
a_t^* \sim \mathrm{Categorical}\left( P(a^0), P(a^1), \dots, P(a^{k-1}) \right)
\label{equation:Categorical}
\end{equation}
During deployment, the optimal action is selected via deterministic execution:
\begin{equation}
a_t^* = \underset{i \in \{0,\dots,k-1\}}{\mathrm{argmax}}  P\left(a^i\right)
\end{equation}

\subsection{Online Reinforcement Learning}
\label{subsection: Online Reinforcement Learning}
The action selector is trained via online RL using sparse rewards. To minimize the burden of manual reward design, we provide a positive reward signal only upon successful completion of a task or subtask. We adopt Proximal Policy Optimization (PPO)\cite{schulman2017proximal} as the underlying RL algorithm, sampling actions from the categorical distribution output by the selector network. Unlike direct RL fine-tuning of the base policy, TAS freezes the parameters of the base policy during training, preventing catastrophic forgetting of the pre-trained capabilities.

In robotic manipulation, motion smoothness is often desirable for both task success and hardware longevity. Leveraging the inherent continuity between adjacent actions within the same chunk, we introduce a coherence penalty during training, defined as follows:
\begin{equation}
\begin{aligned}
r_{\mathrm{coh\_penalty}} &= -\lambda \left\| a_t^* - \mathrm{succ}(a_{t-1}^*) \right\|^2_2 \\
r_{\mathrm{total}} &= r_{\mathrm{task}} + r_{\mathrm{coh\_penalty}}
\end{aligned}
\end{equation}
where $\lambda$ modulates the penalty intensity, successor action $\mathrm{succ}(a_{t-1}^*)$ denotes the immediately following action in the same action chunk as $a_{t-1}^*$, and $r_{\mathrm{task}}$ denotes the sparse reward received only upon task completion. 

By penalizing large deviations between the currently selected action and the natural continuation of the previously selected action, this coherence penalty promotes smoother motion trajectories and contributes to improved stability of the learned policy during deployment.

\subsection{Integration with Residual Reinforcement Learning}
\label{subsection: Integration with Residual Reinforcement Learning}

\begin{figure}[t]
    \centering
    \includegraphics[width=0.43\textwidth]{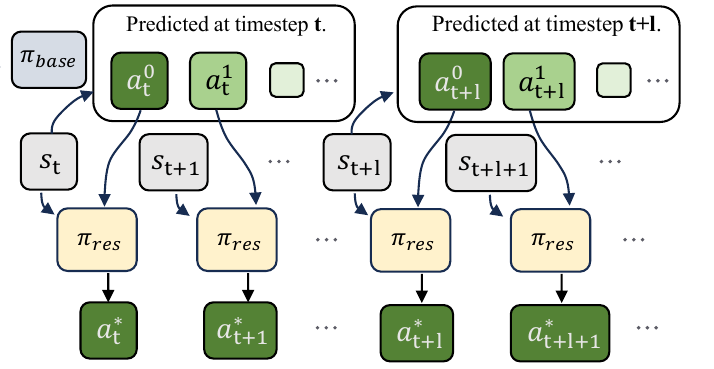}
    \caption{Standard residual RL architecture.}
    \label{fig:residual-baseline}
\end{figure}

\begin{figure}[t]
    \centering
    \includegraphics[width=0.42\textwidth]{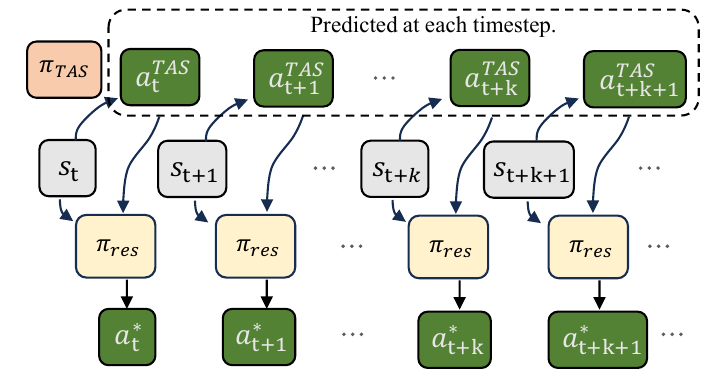}
    \caption{TAS-integrated residual RL architecture.}
    \label{fig:tas-residual}
\end{figure}
% TODO：补充两图的说明文字

Residual RL is a widely adopted paradigm for refining a base policy by learning an additive corrective term. Its core formulation is given by:
\begin{equation}
a_t^* = a_t^{\text{base}} + a_t^{\text{res}}
\end{equation}
where $a_t^{\text{base}}$ denotes the action output by the frozen base policy $\pi_{\text{base}}$, and $a_t^{\text{res}}$ is the residual correction produced by a separately learned residual policy. An inherent limitation of this approach is that the final performance is fundamentally bounded by the quality of the underlying base policy: a weak or suboptimal base policy can severely constrain the efficacy of the residual component.

Since TAS significantly enhances base policy performance, it naturally provides a stronger foundation upon which residual RL can build. This leads to a modified formulation:
\begin{equation}
a_t^* = a_t^{\text{TAS}} + a_t^{\text{res}}
\end{equation}
where $a_t^{\text{TAS}}$ denotes the action selected by the TAS module. The architectural distinction between conventional residual RL and the proposed TAS-integrated variant is illustrated in Fig.~\ref{fig:residual-baseline} and Fig.~\ref{fig:tas-residual}. In the conventional setup, the base policy typically generates a new action chunk only once every $l$ timesteps, leaving the residual policy to provide corrective adjustments based on outdated base actions during the intervening steps. In contrast, the TAS-integrated pipeline performs action selection at every timestep by dynamically choosing the most suitable candidate from $\mathcal{A}_t$. Consequently, the residual policy operates on a foundation that is continuously refreshed with the most contextually appropriate action derived from the latest available observations, thereby enhancing overall performance and reducing the corrective burden placed on the residual learner.

We investigate two concrete integration methodologies that differ in how the TAS module is treated during the residual learning phase:
\begin{itemize}
    \item \textbf{TAS (Frozen) + Residual RL}: In this configuration, the pre-trained TAS module is kept fixed and serves as a static, optimized base policy. Only the parameters of the residual policy are updated during RL training, preserving the behavior of the TAS selector while allowing the residual component to compensate for any remaining task-specific discrepancies.
    \item \textbf{TAS + Residual RL}: In this configuration, the TAS module and the residual policy are co-optimized simultaneously. This allows the selector network to adapt its candidate preferences in response to the corrections introduced by the residual stream, potentially unlocking synergistic improvements beyond what either component could achieve in isolation.
\end{itemize}

\section{Experiments}

We conducted comprehensive experiments in simulation and on hardware to address the following research questions:
\begin{enumerate}
    \item How does TAS perform across diverse tasks and base policy architectures?
    \item How do the distinct action selector architectures compare in performance?
    \item What is the effect of the coherence penalty on TAS?
    \item Is caching actions derived from observations at distinct timesteps essential for TAS performance?
    \item Does TAS-residual RL integration enhance training efficiency and performance?
\end{enumerate}

\subsection{Tasks, Datasets, and Base Policies}
\label{subsection: Tasks, Datasets, and Base Policies}

\begin{figure}[t]
    \centering
    \subfloat[]{\includegraphics[width=0.22\textwidth]{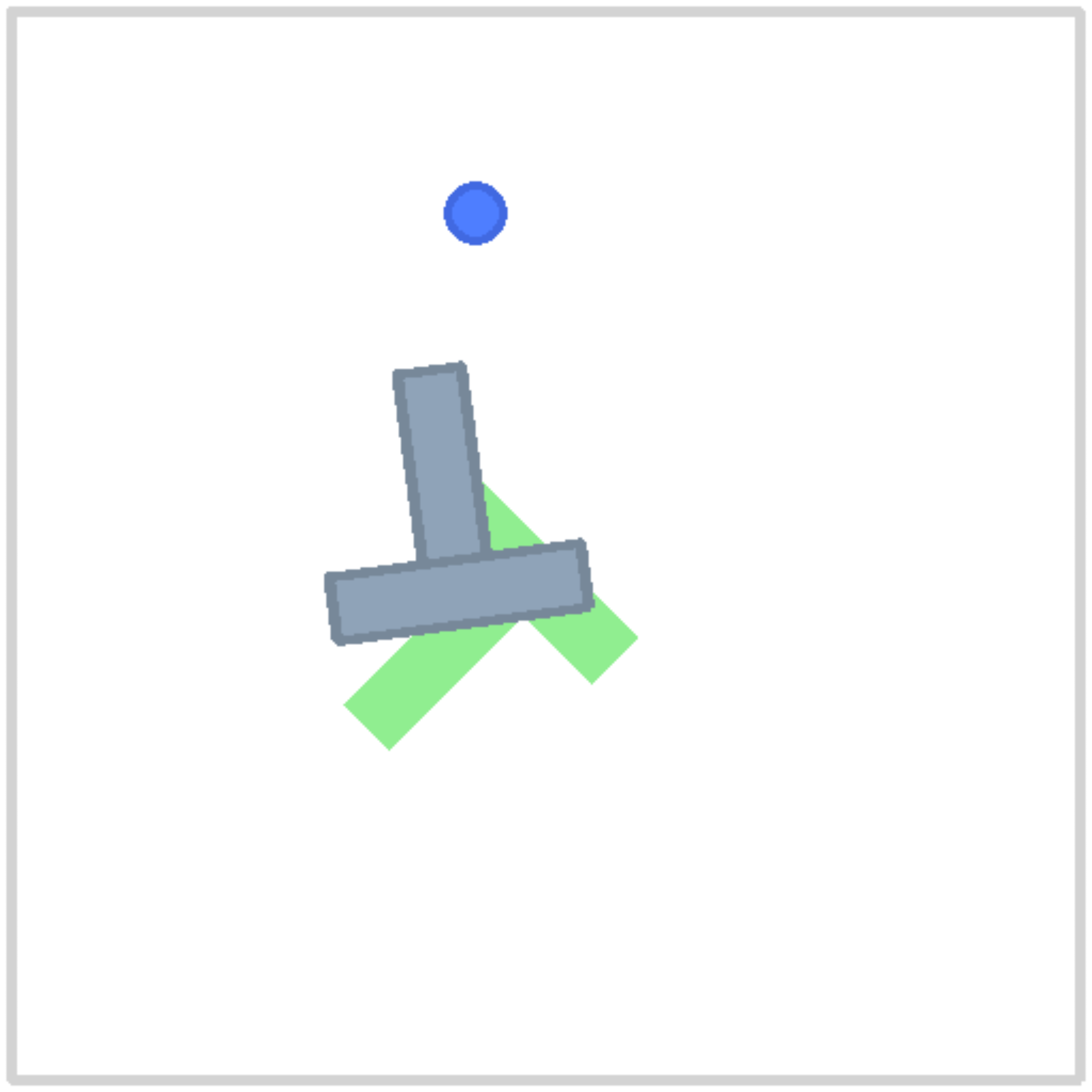}\label{fig:fig1}}
    \hspace{0.5em}
    \subfloat[]{\includegraphics[width=0.22\textwidth]{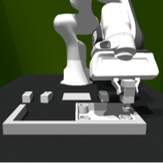}\label{fig:fig2}}
    
    \subfloat[]{\includegraphics[width=0.22\textwidth]{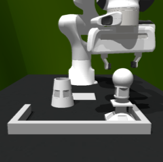}\label{fig:fig3}}
    \hspace{0.5em}
    \subfloat[]{\includegraphics[width=0.22\textwidth]{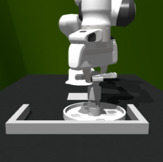}\label{fig:fig4}}
    
    \caption{Simulated environments: (a) PushT\cite{chi2023diffusionpolicy}; 
             (b) One Leg,  (c) Lamp, and  (d) Round Table from FurnitureBench\cite{heo2023furniturebench}.}
    \label{fig:simulated_environments}  % 注意原标签中有空格，已改为下划线
\end{figure}

We evaluated TAS and conducted ablation studies on PushT, Image PushT, and FurnitureBench task suite, which are visualized in Fig.~\ref{fig:simulated_environments}.

\textbf{PushT}: The observation space consists of the 2D positions of nine keypoints on the T-shaped block together with the end-effector pose. We use the 206 human demonstrations released in \cite{chi2023diffusionpolicy}. Two metrics are reported: success rate (SR), defined as achieving a coverage $\geqslant 0.95$ between the T-block and the target pose, and the average of per-episode maximum coverage (MS), a commonly used metric for this environment. All task parameters, including the maximum number of steps per episode, follow the settings in the aforementioned reference.

\textbf{Image PushT}: This variant differs only in the observation space, which consists of an RGB image alongside the end-effector position. All other settings, including the dataset and evaluation protocol, are kept consistent with the standard PushT task.

\textbf{FurnitureBench}: We select three tasks from the FurnitureBench\cite{heo2023furniturebench},\cite{jiang2024transic} task suite: One Leg, Lamp, and Round Table. The observation space comprises the pose information of all object parts and the proprioceptive state of the robot. We utilize the dataset provided in \cite{ankile2024imitationrefinementresidual}, which contains 50 human demonstrations for each task. The success rate (SR) is used as the primary evaluation metric. For the Lamp and Round Table tasks, which consist of two sequential subtasks, we report both SR\textsubscript{1} (subtask-1 success rate) and SR\textsubscript{2} (subtask-2 success rate, which corresponds to the overall task success rate). All termination conditions and the maximum number of steps per episode adhere to the specifications in the same reference.

To emulate real-world sensorimotor uncertainty and facilitate simulation-to-real transfer, we inject two forms of noise into the One Leg and Lamp tasks during both training and evaluation:
\begin{itemize}
  \item \textbf{Per-episode systematic bias}: Fixed random offset per episode $\sim \mathcal{N}(0, \Sigma_{\text{sys}})$.
  \item \textbf{Per-timestep observation noise}: $\mathcal{N}(0, \Sigma_{\text{obs}})$ added at each timestep.
\end{itemize}
The covariance matrices $\Sigma_{\text{sys}}$ and $\Sigma_{\text{obs}}$ were derived from measurements of a pose estimator\cite{wen2024foundationpose} deployed in simulation. Specific noise parameters are provided in Table \ref{tab:noise_params}.

Additionally, we deployed the trained One Leg policy in a real-world setting. Using pose estimation from Intel RealSense D435 cameras\cite{wen2024foundationpose}, policies trained entirely in simulation were transferred directly to the physical robot without further tuning.

Four base policies were evaluated: multi-layer perceptron (MLP), conditional variational autoencoder (CVAE), VQ-BeT, and Diffusion Policy. For the PushT and Image PushT tasks, the implementations of VQ-BeT and Diffusion Policy follow the official code releases from their respective original works\cite{chi2023diffusionpolicy},\cite{lee2024behavior}, while the MLP and CVAE policies were implemented by us. Since ACT\cite{zhao2023learning} does not provide a low-dimensional variant and exhibits limited performance on the Image PushT task, we incorporate only its CVAE backbone and action chunking structure for the PushT task, with an MLP serving as both encoder and decoder. For the FurnitureBench tasks, the MLP and Diffusion Policy architectures follow the setup described in \cite{ankile2024imitationrefinementresidual}.

% \begin{figure}[t]
%     \centering
%     \subfloat[]{\includegraphics[width=0.115\textwidth]{task1.png}\label{fig:fig1}}
%     \hfill
%     \subfloat[]{\includegraphics[width=0.115\textwidth]{task2.png}\label{fig:fig2}}
%     \hfill
%     \subfloat[]{\includegraphics[width=0.115\textwidth]{task3.png}\label{fig:fig3}}
%     \hfill
%     \subfloat[]{\includegraphics[width=0.115\textwidth]{task4.png}\label{fig:fig4}}
%     \caption{Simulated environments: (a) PushT\cite{chi2023diffusionpolicy}; 
%              (b-d) One Leg, Lamp, and Round Table from FurnitureBench\cite{heo2023furniturebench}.}
%     \label{fig:simulated environments}
% \end{figure}

\begin{table}[t]
  \centering
  \caption{Noise parameters for FurnitureBench.}
  \label{tab:noise_params}
  \setlength{\tabcolsep}{4pt}
  \renewcommand{\arraystretch}{1.4}
  \begin{tabular}{l c c c}
    \hline
    Parameter & Distribution & Type & Unit \\
    \hline
    Parts Position & $\mathcal{N}(0, (0.002)^2)$ & Systematic & m \\
    Parts Position & $\mathcal{N}(0, (0.002)^2)$ & Observation & m \\
    Parts Rotation & $\mathcal{N}(0, (0.02)^2)$ & Systematic & rad \\
    Parts Rotation & $\mathcal{N}(0, (0.02)^2)$ & Observation & rad \\
    EE Position & $\mathcal{N}(0, (0.002)^2)$ & Observation & m \\
    EE Rotation & $\mathcal{N}(0, (0.02)^2)$ & Observation & rad \\
    EE Linear Velocity & $\mathcal{N}(0, (0.01)^2)$ & Observation & m/s \\
    EE Angular Velocity & $\mathcal{N}(0, (0.01)^2)$ & Observation & rad/s \\
    \hline
  \end{tabular}
\end{table}

\begin{table*}[htbp]
    \centering
    \caption{Comparison on PushT.}
    \label{tab:Comparison on PushT}
    \renewcommand{\arraystretch}{1.03}
    \begingroup
    \setlength{\tabcolsep}{1.8pt}
    \begin{tabular}{
        l
        >{\centering\arraybackslash}p{0.90cm}
        >{\centering\arraybackslash}p{1.05cm}
        >{\centering\arraybackslash}p{0.90cm}
        >{\centering\arraybackslash}p{1.05cm}
        >{\centering\arraybackslash}p{0.90cm}
        >{\centering\arraybackslash}p{1.05cm}
        >{\centering\arraybackslash}p{0.90cm}
        >{\centering\arraybackslash}p{1.05cm}
        >{\centering\arraybackslash}p{0.90cm}
        >{\centering\arraybackslash}p{1.05cm}
        >{\centering\arraybackslash}p{0.90cm}
        >{\centering\arraybackslash}p{1.05cm}
        >{\centering\arraybackslash}p{0.90cm}
        >{\centering\arraybackslash}p{1.05cm}
    }
    \toprule
    & \multicolumn{8}{c}{\textbf{PushT}} & \multicolumn{6}{c}{\textbf{Image PushT}} \\
    \cmidrule(lr){2-9} \cmidrule(lr){10-15}
    & \multicolumn{2}{c}{\textbf{MLP}} & \multicolumn{2}{c}{\textbf{CVAE}} & \multicolumn{2}{c}{\textbf{VQ-BeT}} & \multicolumn{2}{c}{\textbf{DP}} & \multicolumn{2}{c}{\textbf{MLP}} & \multicolumn{2}{c}{\textbf{VQ-BeT}} & \multicolumn{2}{c}{\textbf{DP}} \\
    \cmidrule(lr){2-3} \cmidrule(lr){4-5} \cmidrule(lr){6-7} \cmidrule(lr){8-9} \cmidrule(lr){10-11} \cmidrule(lr){12-13} \cmidrule(lr){14-15}
    & \textbf{SR} & \textbf{MS} & \textbf{SR} & \textbf{MS} & \textbf{SR} & \textbf{MS} & \textbf{SR} & \textbf{MS} & \textbf{SR} & \textbf{MS} & \textbf{SR} & \textbf{MS} & \textbf{SR} & \textbf{MS} \\
    \midrule
    Vanilla-O
    & 33.0\%  & 0.662  & 39.5\%  & 0.776 & 46.0\%  & 0.704 & 72.6\%  & 0.938 & 32.0\%  & 0.703  & 40.2\%  & 0.613 & 64.9\%  & 0.899 \\
    Vanilla-C
    & 0.2\%  & 0.222  & 3.7\%  & 0.361 & 30.2\%  & 0.560 & 64.9\%  & 0.884 & 9.7\%  & 0.422  & 12.5\%  & 0.364 & 48.3\%  & 0.796 \\
    EMA\cite{zhao2023learning}  & 33.2\%  & 0.678 & 43.8\%  & 0.780   & 42.6\%  & 0.651 & 74.1\%  & 0.919 & 29.7\%  & 0.659 & 37.1\%  & 0.581 & 63.6\%  & 0.858 \\
    BID-O\cite{liu2025bidirectional}    & /  & / & 43.1\%  & 0.793   & 43.4\%  & 0.676 & 73.3\%  & 0.934 & /  & /  & 45.0\%  & 0.696 & 65.0\%  & 0.931 \\
    BID-C\cite{liu2025bidirectional}   & /  & / & 5.5\%  & 0.407   & 23.1\%  & 0.487 & 65.8\%  & 0.877 & /  & /  & 7.0\%  & 0.374 & 63.0\%  & 0.789 \\
    TAS (ours)       & \textbf{88.2\%}  & \textbf{0.921} & \textbf{95.3\%}  & \textbf{0.970} & \textbf{78.1\%}  & \textbf{0.829} & \textbf{96.0\%}  & \textbf{0.980} & \textbf{82.8\%}  & \textbf{0.892} & \textbf{72.3\%}  & \textbf{0.802} & \textbf{89.9\%}  & \textbf{0.938}  \\
    FineTune         & 2.1\%  & 0.792 & 2.3\%  & 0.861   & /  & /  & /  & / & 15.6\%  & 0.800 & /  & /  & /  & / \\
    \bottomrule
    \end{tabular}
    \endgroup
\end{table*}

\begin{table*}[htbp]
    \centering
    \caption{Comparison on FurnitureBench (Noise-Free Tasks).}
    \label{tab:Comparison on FurnitureBench (a)}
    \begin{tabular}{l *{10}{>{\centering\arraybackslash}p{1.1cm}}}
    \toprule
        & \multicolumn{2}{c}{\textbf{One Leg}}
        & \multicolumn{4}{c}{\textbf{Lamp}}
        & \multicolumn{4}{c}{\textbf{Round Table}} \\
    \cmidrule(lr){2-3} \cmidrule(lr){4-7} \cmidrule(lr){8-11}
        & \textbf{MLP} & \textbf{DP}
        & \multicolumn{2}{c}{\textbf{MLP}} & \multicolumn{2}{c}{\textbf{DP}}
        & \multicolumn{2}{c}{\textbf{MLP}} & \multicolumn{2}{c}{\textbf{DP}} \\
    \cmidrule(lr){2-2} \cmidrule(lr){3-3} 
    \cmidrule(lr){4-5} \cmidrule(lr){6-7}
    \cmidrule(lr){8-9} \cmidrule(lr){10-11}
        & \textbf{SR} & \textbf{SR}
        & \textbf{SR\textsubscript{1}} & \textbf{SR\textsubscript{2}}  & \textbf{SR\textsubscript{1}} & \textbf{SR\textsubscript{2}}
        & \textbf{SR\textsubscript{1}} & \textbf{SR\textsubscript{2}}  & \textbf{SR\textsubscript{1}} & \textbf{SR\textsubscript{2}} \\
    \midrule
    Vanilla-O
        & 50.2\% & 63.8\% 
        & 19.6\% & 8.2\% & 11.6\% & 7.2\%
        & 74.6\% & 3.8\% & 60.3\% & 5.4\% \\
    Vanilla-C
        & 0.0\% & 0.0\% 
        & 0.0\% & 0.0\% & 0.0\% & 0.0\%
        & 0.0\% & 0.0\% & 0.0\% & 0.0\% \\
    EMA\cite{zhao2023learning}
        & 43.6\% & 53.0\% 
        & 10.5\% & 5.0\% & 10.5\% & 4.5\%
        & 74.2\% & 0.7\% & 62.9\% & 4.2\% \\
    BID-O\cite{liu2025bidirectional}
        & / & 0.0\% 
        & / & / & 0.0\% & 0.0\%
        & / & / & 0.0\% & 0.0\% \\
    BID-C\cite{liu2025bidirectional}
        & / & 62.4\% 
        & / & / & 11.5\% & 5.4\%
        & / & / & 71.5\% & 4.0\% \\
    TAS (ours)
        & 74.4\% & \textbf{78.1\%} 
        & \textbf{67.8\%} & \textbf{55.2\%} & \textbf{61.4\%} & \textbf{51.6\%}
        & \textbf{96.6\%} & \textbf{77.1\%} & \textbf{92.5\%} & \textbf{54.7\%} \\
    FineTune
        & \textbf{83.7\%} & / 
        & 60.8\% & 44.7\% & / & /
        & 77.4\% & 7.5\% & / & / \\
    \bottomrule
    \end{tabular}
\end{table*}

% 有噪声任务表格
\begin{table}[htbp]
    \centering
    \caption{Comparison on FurnitureBench (Noisy Tasks).}
    \label{tab:Comparison on FurnitureBench (b)}
    \begin{tabular}{l *{6}{>{\centering\arraybackslash}p{0.72cm}}}
    \toprule
        & \multicolumn{2}{c}{\textbf{One Leg Noise}}
        & \multicolumn{4}{c}{\textbf{Lamp Noise}} \\
    \cmidrule(lr){2-3} \cmidrule(lr){4-7}
        & \textbf{MLP} & \textbf{DP} 
        & \multicolumn{2}{c}{\textbf{MLP}} & \multicolumn{2}{c}{\textbf{DP}} \\
    \cmidrule(lr){2-2} \cmidrule(lr){3-3} 
    \cmidrule(lr){4-5} \cmidrule(lr){6-7}
        & \textbf{SR} & \textbf{SR}
        & \textbf{SR\textsubscript{1}} & \textbf{SR\textsubscript{2}}  & \textbf{SR\textsubscript{1}} & \textbf{SR\textsubscript{2}}\\
    \midrule
    Vanilla-O
        & 22.8\% & 14.1\%
        & 18.2\% & 5.0\% & 7.5\% & 2.0\% \\
    Vanilla-C
        & 0.0\% & 0.0\%
        & 0.0\% & 0.0\% & 0.0\% & 0.0\% \\
    EMA\cite{zhao2023learning}
        & 22.7\% & 14.9\%
        & 10.7\% & 4.8\% & 7.9\% & 2.1\% \\
    BID-O\cite{liu2025bidirectional}
        & / & 0.0\%
        & / & / & 0.0\% & 0.0\% \\
    BID-C\cite{liu2025bidirectional}
        & / & 18.9\%
        & / & / & 9.1\% & 2.9\% \\
    TAS (ours)
        & \textbf{68.4\%} & \textbf{63.4\%} 
        & \textbf{61.7\%} & \textbf{51.2\%} & \textbf{46.3\%} & \textbf{36.7\%} \\
    FineTune
        & 37.5\% & / 
        & 20.0\% & 6.8\% & / & / \\
    \bottomrule
    \end{tabular}
\end{table}

\subsection{Performance of TAS}
\label{subsection: Performance of TAS}
To evaluate the performance gains achieved by integrating TAS into base policies, we conducted experiments across a diverse set of tasks and base policy architectures. Under consistent experimental settings, we compared TAS against the following baselines:
\begin{itemize}
    \item \textbf{Vanilla Open-Loop}: Execute the first 8 actions of each predicted action chunk without re-planning.
    \item \textbf{Vanilla Closed-Loop}: Execute only the first action of each chunk and re-plan at every timestep.
    \item \textbf{EMA}\cite{zhao2023learning}: At each timestep, compute the EMA between the current sampled action and historical actions.
    \item \textbf{BID Open-Loop}\cite{liu2025bidirectional}: Sample 20 candidate action chunks every 8 timesteps, select the optimal chunk using the BID scoring criterion, and execute the entire selected chunk.
    \item \textbf{BID Closed-Loop}\cite{liu2025bidirectional}: Sample 20 candidate action chunks at each timestep, select the optimal chunk using BID, and execute only the first action of the selected chunk.
    \item \textbf{RL FineTune}: Fine-tune the pre-trained base policy using PPO with sparse task rewards.
\end{itemize}

Note that BID relies on repeated sampling from the same observation to generate multiple candidates, which is incompatible with deterministic architectures such as the MLP policy. Therefore, BID baselines are omitted for the MLP architectures. Similarly, RL FineTune experiments were conducted exclusively on the MLP and CVAE architectures, as the approach cannot be straightforwardly applied to the VQ-BeT and Diffusion Policy architectures within our experimental framework.

Key hyperparameters were configured as follows. For the MLP, CVAE, and Diffusion Policy architectures, the prediction horizon was set to 32. For the VQ-BeT architecture, the horizon was reduced to 8 to prevent performance degradation of the VQ-VAE when handling high-dimensional outputs. The number of cached action candidates $k$ was set to 8 for the former three policies and to 6 for VQ-BeT. While larger values of $k$ can yield further performance improvements, they also incur additional computational overhead during training. TAS was trained using PPO with 256 parallel environments for the PushT tasks and 1024 for the FurnitureBench tasks. The policy update schedule follows the protocol in \cite{ankile2024imitationrefinementresidual}: each update iteration is performed only after all parallel sub-environments (with auto-reset disabled) have completed one full episode, and the collected episode data from all environments are used in that iteration. Training consisted of 1,000 such update iterations.

The results, summarized in Table~\ref{tab:Comparison on PushT}, Table~\ref{tab:Comparison on FurnitureBench (a)}, and Table~\ref{tab:Comparison on FurnitureBench (b)}, demonstrate that TAS consistently improves performance across all tasks and base policies. On average, TAS yields success-rate gains of 42.58\% on PushT, 35.97\% on Image PushT, and 41.18\% on the FurnitureBench tasks. Comparatively, EMA and BID provide only marginal improvements, and even exhibit performance degradation on certain tasks such as One Leg.

RL FineTune exhibits unstable performance patterns: it achieves the highest success rates in the One Leg task and delivers noticeable improvements in the Lamp task. However, in specific challenging scenarios—noisy environments (e.g., One Leg Noise, Lamp Noise) and tasks with low initial success rates (e.g., Round Table)—RL FineTune suffers from catastrophic forgetting. Its success rates collapse to near-zero levels after initial training phases. For the PushT and Image PushT tasks, RL FineTune improves the average maximum coverage (MS) but fails to achieve meaningful success rates (approaching 0\%). This discrepancy can be explained by the task's two-phase nature: when the T block is far from the target, coarse pushing suffices to increase coverage, a phase where RL fine-tuning is effective, leading to the observed MS gains. However, final success requires high-precision, continuous alignment when the block and target are close. Minute action errors at this stage cause coverage to drop precipitously. Fine-tuning an action-chunking policy with RL lacks the capacity for fine-grained, per-timestep reactivity within the chunk, preventing the necessary precise corrections and leading to near-zero success rates.

\begin{table*}[htbp]
    \centering
    \caption{Comparison of different selector architectures}
    \label{tab:Different selector architectures}
    % \small
    \begin{tabular}{
        l
        >{\centering\arraybackslash}p{1.1cm}
        >{\centering\arraybackslash}p{1.5cm}
        >{\centering\arraybackslash}p{1.1cm}
        >{\centering\arraybackslash}p{1.5cm}
        >{\centering\arraybackslash}p{2.1cm}
        >{\centering\arraybackslash}p{2.1cm}
    }
        \toprule
            & \multicolumn{2}{c}{\textbf{PushT+MLP}} 
            & \multicolumn{2}{c}{\textbf{PushT+CVAE}} 
            & \textbf{OnelegDR+MLP}
            & \textbf{OnelegDR+Diff} \\
        \cmidrule(lr){2-3} \cmidrule(lr){4-5} \cmidrule(lr){6-6} \cmidrule(lr){7-7}
            & \textbf{SR} & \textbf{MS} & \textbf{SR} & \textbf{MS} & \textbf{SR} & \textbf{SR} \\
        \midrule
        Implicit Space-aware Selector (ours)
        & \textbf{88.2\%} & \textbf{0.9205} & \textbf{95.3\%} & \textbf{0.9695} 
        & \textbf{68.4\%} & \textbf{63.4\%} \\
        Implicit MLP Selector 
        & 76.9\% & 0.8631 & 90.8\% & 0.9535 & 55.8\% & 53.0\% \\
        Explicit MLP Selector 
        & 58.5\% & 0.7772 & 71.8\% & 0.9225 & 46.2\% & 40.1\% \\
        \bottomrule
    \end{tabular}
\end{table*}

% \begin{table}[t]
%     \centering
%     \caption{Comparison of different selector architectures}
%     \label{tab:Different selector architectures}
%     \small
%     \setlength{\tabcolsep}{4pt}
%     \begin{tabular}{@{}lcccccc@{}}
%         \toprule
%         & \multicolumn{4}{c}{\textbf{PushT}} & \multicolumn{2}{c}{\textbf{OneLegNoise}} \\
%         \cmidrule(lr){2-5} \cmidrule(lr){6-7}
%         & \multicolumn{2}{c}{\textbf{MLP}} & \multicolumn{2}{c}{\textbf{CVAE}} & \multicolumn{1}{c}{\textbf{MLP}} & \multicolumn{1}{c}{\textbf{DP}} \\
%         \cmidrule(lr){2-3} \cmidrule(lr){4-5} \cmidrule(lr){6-6} \cmidrule(lr){7-7}
%          & \textbf{SR} & \textbf{MS} & \textbf{SR} & \textbf{MS} & \textbf{SR} & \textbf{SR} \\
%         \midrule
%         Space-Aware
%         & \textbf{88.2\%} & \textbf{0.921} & \textbf{95.3\%} & \textbf{0.970} 
%         & \textbf{68.4\%} & \textbf{63.4\%} \\
%         Implicit MLP 
%         & 76.9\% & 0.863 & 90.8\% & 0.954 & 55.8\% & 53.0\% \\
%         Explicit MLP 
%         & 58.5\% & 0.777 & 71.8\% & 0.923 & 46.2\% & 40.1\% \\
%         \bottomrule
%     \end{tabular}
% \end{table}

\subsection{Comparative Analysis of Selector Network Architectures}
\label{subsection: Different Selector Architectures}
We conducted comparative experiments among three selector architectures:
\begin{itemize}
    \item \textbf{Implicit Space-Aware Selector}: The proposed architecture described in Section~\ref{subsection: Temporal Action Selection Architecture}, which computes action scores via cosine similarity between learned embeddings of the task context and each candidate action.
    \item \textbf{Implicit MLP Selector}: For each candidate $a^i \in \mathcal{A}_t$, an MLP takes as input the concatenated vector $[s_t; \mathcal{A}_t; a^i]$ and outputs a scalar score. The scores for all candidates are then normalized via softmax to yield a probability distribution over the candidate set.
    \item \textbf{Explicit MLP Selector}: Following standard architectures for discrete action-space deep reinforcement learning\cite{mnih2015human}, an MLP processes the concatenation of the current observation $s_t$ and the candidate set $\mathcal{A}_t$, and directly outputs a probability distribution over all candidates without per-candidate scoring.
\end{itemize}
The results, presented in Table~\ref{tab:Different selector architectures}, demonstrate that the Implicit Space-Aware Selector consistently achieves superior performance across all four experimental configurations. We attribute this advantage to the structured inductive bias introduced by formulating action scoring as a vector similarity matching problem in a shared latent space. In contrast to the MLP-based selectors—which operate as black-box function approximators—the Space-Aware design reduces the risk of overfitting to individual candidate actions and exhibits improved generalization to previously unseen yet semantically similar actions. Furthermore, the use of cosine similarity inherently bounds the action scores within a normalized range. When combined with temperature scaling, this property leads to more stable exploration behavior compared to the unconstrained score outputs typical of MLP-based approaches.

\begin{table}[t]
  \centering
  \caption{Coherence penalty coefficient ablation.}
  \label{tab:Coherence Penalty Coefficient Ablation}
  \setlength{\tabcolsep}{6pt}
  \renewcommand{\arraystretch}{1.4}
  \begin{tabular}{l c c c}
    \hline
    \textbf{Policy} & \textbf{SR (\%)} & \textbf{Avg. Vel (m/s)} & \textbf{Avg. Acc (m/s$^2$)} \\
    \hline
    Base Policy & 50.2 & 0.0397 & 0.1021 \\
    TAS w $\lambda$=0 & 68.1 & 0.0421 & 0.1417 \\
    TAS w $\lambda$=0.01 & 74.4 & 0.0415 & 0.1253 \\
    TAS w $\lambda$=0.05 & 67.5 & 0.0394 & 0.1058 \\
    \hline
  \end{tabular}
\end{table}

\subsection{Effect of the Coherence Penalty}
\label{subsection: Effect of the Coherence Penalty}
To evaluate the effect of the coherence penalty on policy performance, we conducted a controlled experiment using the MLP base policy on the One Leg task. We systematically varied the penalty intensity $\lambda$ and measured both the task success rate and quantitative metrics of motion smoothness. Since the FurnitureBench environment employs direct Cartesian position control of the end-effector, we adopted the average velocity and average acceleration of the end-effector over the course of an episode as proxies for motion smoothness.

The results are presented in Table~\ref{tab:Coherence Penalty Coefficient Ablation} and indicate a clear trade-off regulated by the penalty intensity. As $\lambda$ increases, motion smoothness improves monotonically, reflected by reductions in both average velocity and acceleration. The task success rate, however, exhibits a non-monotonic trend: it rises under moderate penalty values, reaches a peak, and subsequently declines as the penalty becomes overly dominant. This pattern suggests that a moderate coherence penalty encourages temporally consistent action selection, thereby facilitating smoother trajectories that benefit task execution. Conversely, an excessively large penalty imposes restrictive constraints on policy optimization, limiting the capacity for necessary corrective adjustments and ultimately impairing the final success rate.

\subsection{The Necessity of Action Caching}
\begin{figure}[t]
    \centering
    \includegraphics[width=0.48\textwidth]{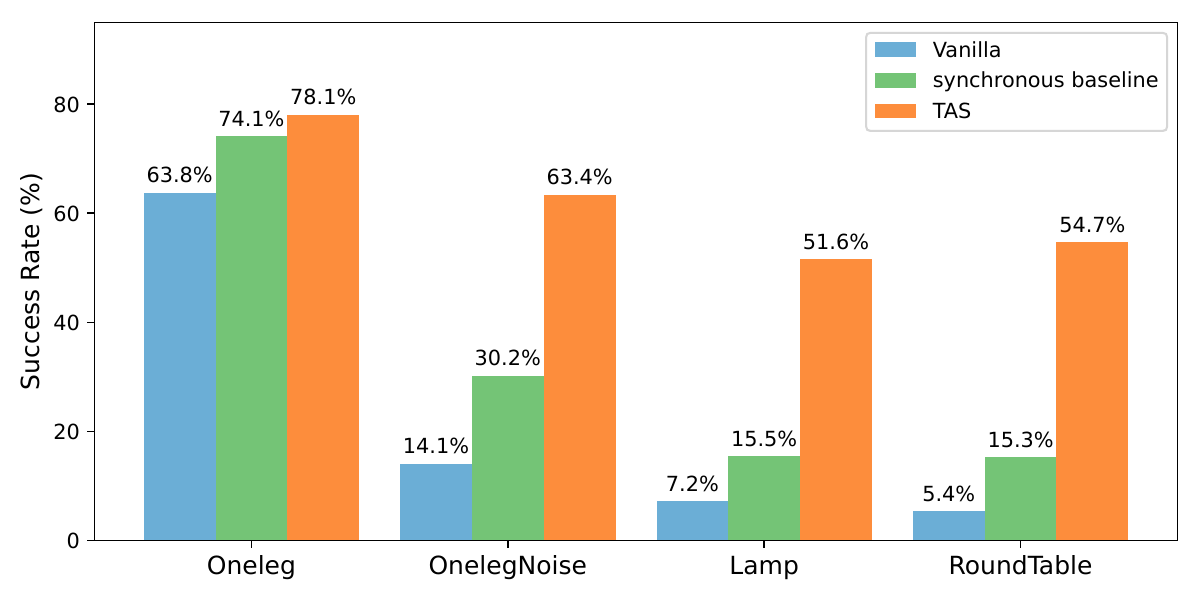}
    \caption{Performance comparison between TAS and the synchronous baseline.}
    \label{fig:Ablation Study on Action Caching}
\end{figure}

Unlike alternative approaches that select optimal actions from multiple candidates generated at the same timestep\cite{liu2025bidirectional}, TAS constructs its candidate set using observations drawn from distinct historical timesteps. This design enables the simultaneous optimization of reactivity and decision consistency. To isolate and validate the contribution of multi-timestep observations, we designed an ablation study that compares TAS against a synchronous baseline. In this baseline, $k$ candidate action chunks are predicted simultaneously at fixed intervals of $k$ timesteps, with all candidates derived from the same observation available at that moment. This contrasts with TAS's mechanism of generating candidates from distinct historical observations. The results, presented in Fig.~\ref{fig:Ablation Study on Action Caching}, indicate that TAS yields substantially higher success rates, particularly in regimes where the vanilla policy performs poorly, thereby confirming the essential role of multi-timestep caching.

To further characterize the dynamic selection behavior of TAS, we analyzed 800 successful rollouts from the One Leg task after aligning their temporal axes using Dynamic Time Warping (DTW)\cite{jablonski2011quaternion}. We quantified two principal metrics: (i) the per-timestep selection probability for each action index, denoted $P_0 \sim P_7$, where $P_0$ corresponds to the most recently generated action and $P_7$ corresponds to the earliest cached action in the candidate set; and (ii) the probabilities of transition patterns between action indices across consecutive timesteps. We categorize these transitions into three distinct patterns:
\begin{itemize}
    \item \textbf{SAME}: The selected action originates from the same action chunk as the action chosen at the previous timestep.
    \item \textbf{NEWER}: The selected action is more recent than that required for a SAME transition.
    \item \textbf{OLDER}: The selected action was predicted earlier than that required for a SAME transition.
\end{itemize}
When the previous timestep employed the final available candidate in its chunk, any subsequent selection is classified as Pattern SAME. The probabilities of these patterns are denoted $P_{\mathrm{SAME}}$, $P_{\mathrm{NEWER}}$, and $P_{\mathrm{OLDER}}$.

\begin{figure*}[htbp]
    \centering
    \includegraphics[width=0.9\textwidth]{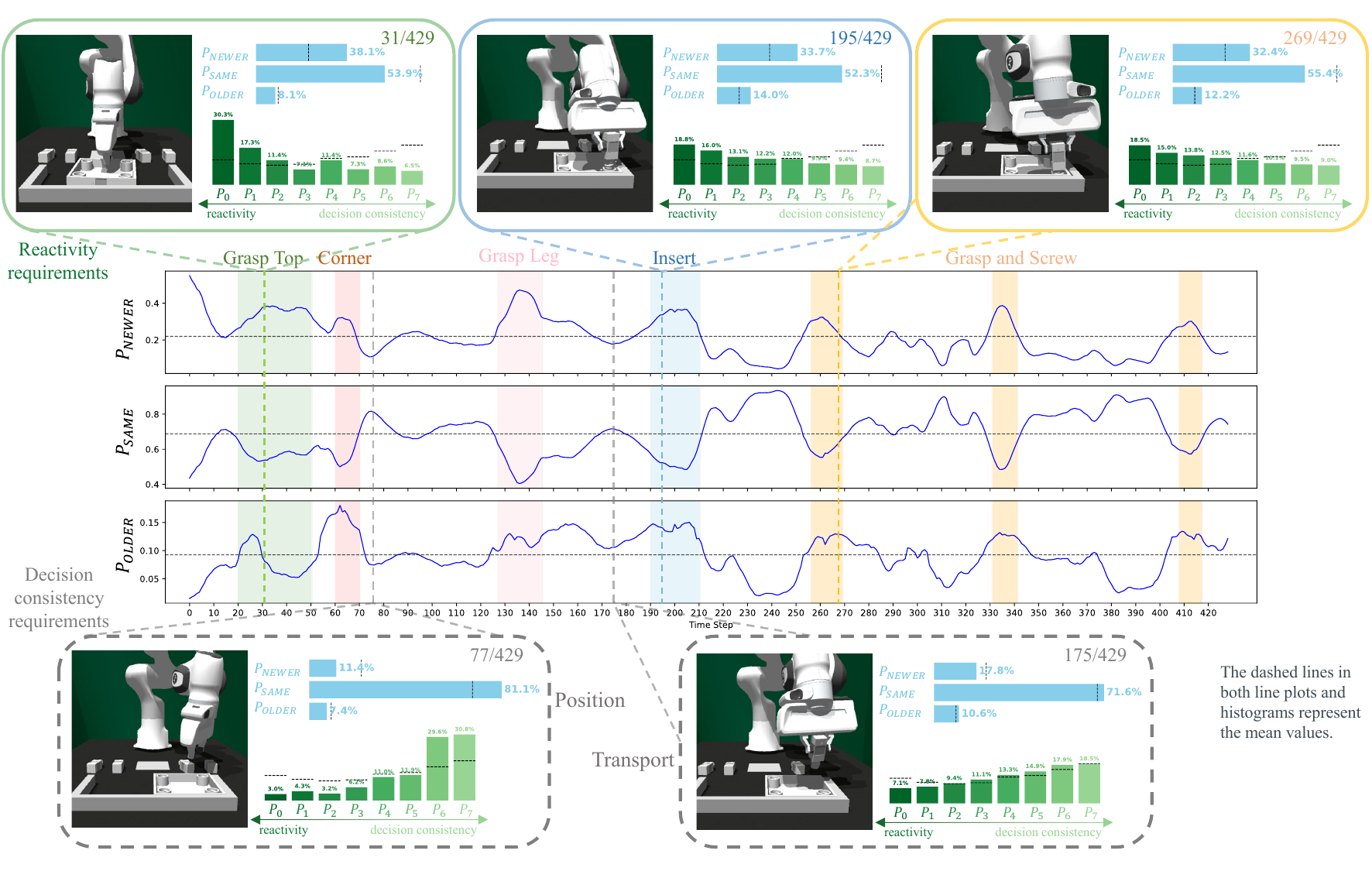}
    \caption{Temporal evolution of TAS selection patterns across task phases.}
    \label{fig:action_probabilities}
\end{figure*}

Fig.~\ref{fig:action_probabilities} illustrates the temporal evolution of these three pattern probabilities. At five representative timesteps, the figure includes visual snapshots alongside histograms of the corresponding index probabilities and transition pattern probabilities. Dashed lines indicate the overall average of each metric. During precision-critical operations (e.g., grasping, corner adjustment, and insertion), TAS exhibits a preference for lower action indices (higher $P_0$--$P_3$) and an elevated probability of NEWER transitions ($P_{\mathrm{NEWER}}$). Conversely, during low-precision phases (e.g., positioning and transport), the policy shifts its preference toward higher indices (higher $P_4$--$P_7$) with a corresponding increase in SAME transitions ($P_{\mathrm{SAME}}$).

\begin{figure}[t]
    \centering
    \includegraphics[width=0.45\textwidth]{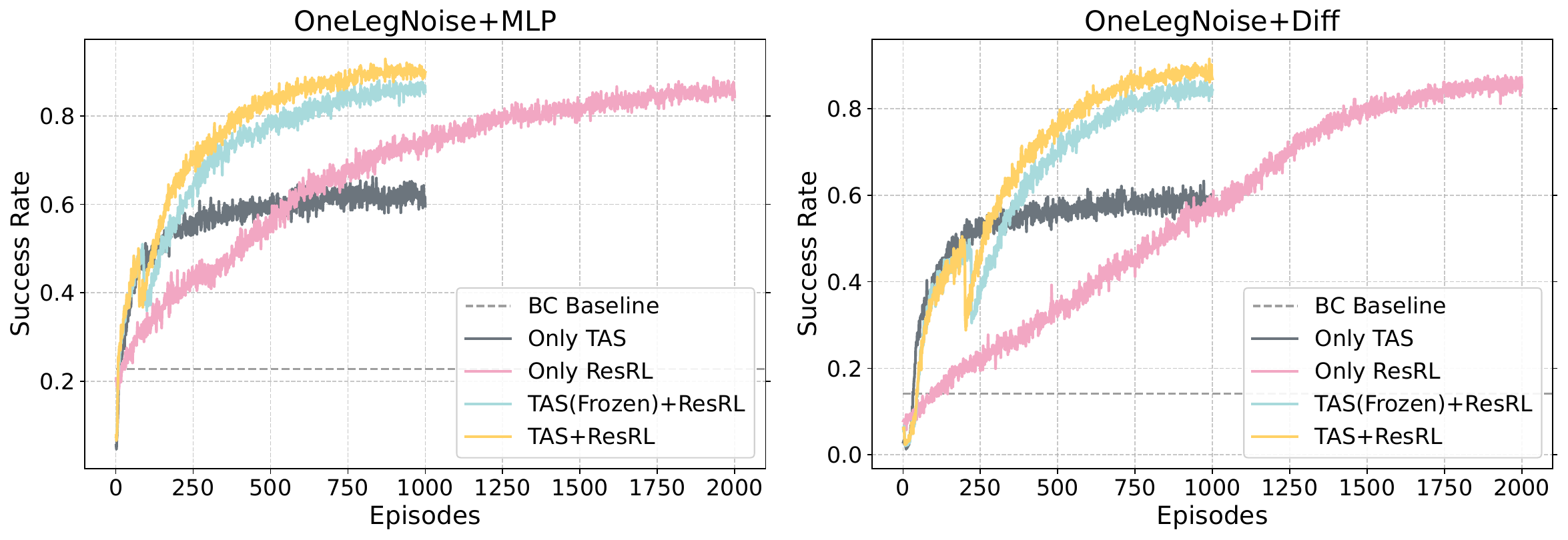}
    \caption{Comparison of TAS-integrated residual RL.}
    \label{fig:Comparison of TAS-integrated residual RL}
\end{figure}

\subsection{Integration with Residual Reinforcement Learning}
\label{subsection:residual RL integration}
In addition to serving as a direct enhancement to base policies, TAS can also function as an improved base policy for residual reinforcement learning or be jointly optimized with a residual component. We conducted comparative experiments across the following five configurations:
\begin{enumerate}
    \item \textit{Base Policy}: The original policy (e.g., MLP or Diffusion Policy) without any modification.
    \item \textit{Only TAS}: The base policy augmented solely with the TAS.
    \item \textit{Only Residual RL}\cite{ankile2024imitationrefinementresidual}: The base policy augmented with residual RL, without TAS.
    \item \textit{TAS (Frozen) + Residual RL}: TAS serves as a fixed, enhanced base policy for residual RL after the initial training phase.
    \item \textit{TAS + Residual RL}: TAS and the residual policy are jointly optimized after the initial training phase.
\end{enumerate}
The residual RL component follows the same architecture as described in \cite{ankile2024imitationrefinementresidual}. For both the TAS (Frozen) + Residual RL and TAS + Residual RL configurations, training proceeds in two distinct stages. In the first stage, only the TAS module is optimized. Once the success rate attains a predefined threshold (e.g., 50\%), the second stage is initiated, during which residual policy training is activated. In the TAS (Frozen) + Residual RL configuration, the TAS module is subsequently frozen; in the TAS + Residual RL configuration, it continues to be updated jointly with the residual policy.

The results are presented in Fig.~\ref{fig:Comparison of TAS-integrated residual RL}. The TAS + Residual RL configuration achieves the highest training efficiency and final performance ceiling among all evaluated variants. Relative to Only TAS, the addition of the residual module enables performance to surpass the inherent limitations of the base policy, yielding a substantial overall gain. Compared with Only Residual RL, the inclusion of TAS markedly accelerates training progress and produces a modest elevation of the asymptotic performance level. Furthermore, the jointly optimized variant outperforms the frozen configuration. A temporary performance decline is observed at the transition between training stages in both TAS (Frozen) + Residual RL and TAS + Residual RL configurations; this effect is attributable to the cold start of the newly introduced residual module. Performance subsequently recovers and rises rapidly as training continues.

\subsection{Real World Experiments}
\label{subsection:Real World Experiments}
\begin{figure*}[t]
    \centering
    \includegraphics[width=0.8\textwidth]{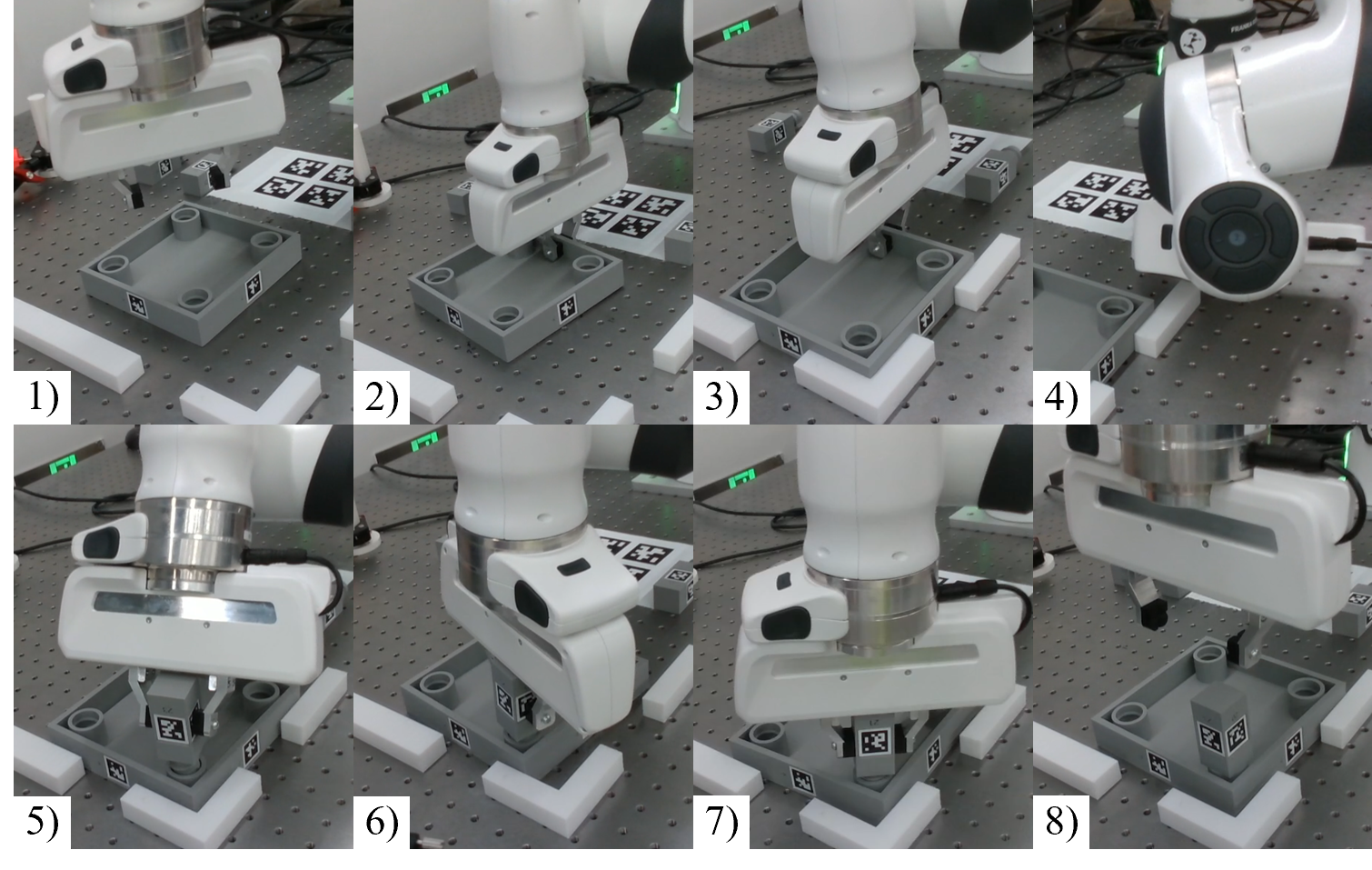}
    \caption{Full workflow execution of the One Leg task in real-world environments.}
    \label{fig:real-world experiment process}
\end{figure*}

\begin{figure*}[t]
    \centering
    \subfloat[]{\includegraphics[width=0.95\textwidth]{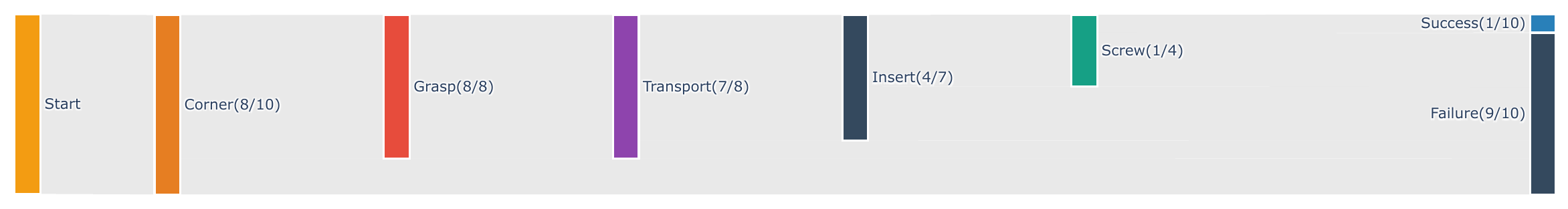}\label{fig:sankey_diagram_2}}
    \\
    \subfloat[]{\includegraphics[width=0.95\textwidth]{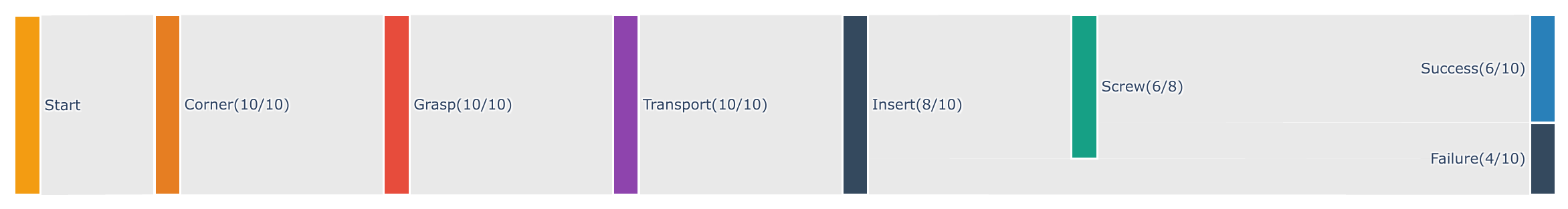}\label{fig:sankey_diagram_1}}
    \caption{Sankey diagrams tracking real-world task flow for (a) Only residual RL and (b) TAS + residual RL, showing the proportion of successful completions at each stage.}
    \label{fig:sankey_diagrams}
\end{figure*}

To further evaluate policy robustness, we deployed the policies trained on the One Leg Noise task in a real-world setting, as illustrated in Fig.~\ref{fig:real-world experiment process}. Since the Base Policy and Only TAS configurations did not achieve reliable task completion, we report in Fig.~\ref{fig:sankey_diagrams} the stage-wise success rates for the Only Residual RL and TAS + Residual RL configurations. The TAS + Residual RL policy attains a markedly higher overall success rate than Only Residual RL.

This performance discrepancy stems from two principal limitations of the Only Residual RL approach. First, its native action chunking mechanism exhibits susceptibility to sensor noise: an isolated noise spike can induce the generation of a suboptimal action chunk whose effects persist across multiple subsequent timesteps. In contrast, the TAS + Residual RL policy is capable of promptly re-evaluating and correcting such suboptimal actions by leveraging the set of historically cached candidates, thereby limiting error propagation. Second, the Only Residual RL policy displays periodic execution latency that is synchronized with action chunk boundaries, resulting in discernible pauses during task execution. Collectively, these two failure modes contribute to the substantial performance degradation observed in the Only Residual RL policy.

\section{Conclusions and Future Work}
In this work, we proposed Temporal Action Selection, a method designed to address the inherent trade-off between reactivity and decision consistency in action chunking. Experimental results across diverse tasks and base policy architectures demonstrate consistent performance gains achieved by TAS, while a series of ablation studies isolate and validate the contributions of its individual components. Beyond serving as a direct policy enhancement, TAS can also function as an improved base policy for residual RL. When integrated with residual RL, TAS yields improvements in both training efficiency and the asymptotic performance ceiling. This integrated formulation is directly compatible with existing frameworks that employ residual RL to refine action-chunking imitation learning policies.

A current limitation of the TAS training procedure is its reliance on online reinforcement learning. While this choice is motivated by considerations of sample efficiency and the need to avoid potentially damaging interactions with physical robots during exploration, it introduces an unavoidable dependence on simulation environments. Recent progress in offline reinforcement learning\cite{intelligence2025pi},\cite{brandfonbrener2023visual} and real-world online policy optimization\cite{luo2025precise} suggests promising directions for overcoming this constraint. Future work will investigate training paradigms informed by these advances, with the objective of eliminating the reliance on simulation, narrowing the sim-to-real performance gap, and enabling more efficient and robust policy learning directly in real-world settings.

%                                   % the textheight of the last page by a suitable amount.
%                                   % This command does not take effect until the next page
%                                   % so it should come on the page before the last. Make
%                                   % sure that you do not shorten the textheight too much.

%%%%%%%%%%%%%%%%%%%%%%%%%%%%%%%%%%%%%%%%%%%%%%%%%%%%%%%%%%%%%%%%%%%%%%%%%%%%%%%%

%%%%%%%%%%%%%%%%%%%%%%%%%%%%%%%%%%%%%%%%%%%%%%%%%%%%%%%%%%%%%%%%%%%%%%%%%%%%%%%%

%%%%%%%%%%%%%%%%%%%%%%%%%%%%%%%%%%%%%%%%%%%%%%%%%%%%%%%%%%%%%%%%%%%%%%%%%%%%%%%%
% \section*{ACKNOWLEDGMENT}

% The preferred spelling of the word ÒacknowledgmentÓ in America is without an ÒeÓ after the ÒgÓ. Avoid the stilted expression, ÒOne of us (R. B. G.) thanks . . .Ó  Instead, try ÒR. B. G. thanksÓ. Put sponsor acknowledgments in the unnumbered footnote on the first page.

\bibliographystyle{IEEEtran}  % 指定参考文献样式
\bibliography{IEEEabrv, ref}     % 指定引用的 .bib 文件，不加扩展名

\end{document}